\documentclass[sii]{ipart}

\usepackage{mathrsfs}
\usepackage{natbib,graphicx,lscape,longtable}
\usepackage{amsmath,amsthm,amssymb,color}
\usepackage{xcolor}

\usepackage{booktabs}
\usepackage{multirow}
\usepackage{engord}

\RequirePackage{hyperref}

\startlocaldefs
\theoremstyle{plain}

\endlocaldefs

\pubyear{2022}
\volume{0}
\issue{0}
\firstpage{1}
\lastpage{11}

\begin{document}

\begin{frontmatter}
\title{Improved Na{\"i}ve Bayes with Mislabeled Data}

\begin{aug}
    \author{\inits{Q.}\fnms{Qianhan} \snm{Zeng}\ead[label=e1]{helenology@stu.pku.edu.cn}},
    \address{Guanghua School of Management, Peking University\\
    \printead{e1}}
    \author{\inits{Y.}\fnms{Yingqiu} \snm{Zhu}\thanksref{t2}\ead[label=e2]{rozen0maiden@126.com}},
    \address{School of Statistics, University of International Business and Economics\\
    \printead{e2}}
    \author{\inits{X.}\fnms{Xuening} \snm{Zhu}\ead[label=e3]{xueningzhu@fudan.edu.cn}},
    \address{School of Data Science, Fudan University\\
    \printead{e3}}
    \author{\inits{F.}\fnms{Feifei} \snm{Wang}\ead[label=e4]{feifei.wang@ruc.edu.cn}},
    \address{School of Statistics, Renmin University of China\\
    \printead{e4}}
    \author{\inits{W.}\fnms{Weichen} \snm{Zhao}\ead[label=e5]{zhaoweichen@hhu.edu.cn}},
    \address{College of Computer and Information, Hohai University\\
    \printead{e5}}
    \author{\inits{S.}\fnms{Shuning} \snm{Sun}\ead[label=e6]{snsun21@m.fudan.edu.cn}},
    \address{School of Data Science, Fudan University \\
    \printead{e6}}
    \author{\inits{S.}\fnms{Meng} \snm{Su}\ead[label=e7]{meng.su@percent.cn}},
    \address{Beijing Percent Technology Group Co., Ltd.\\
    \printead{e7}}
    \and
    \author{\inits{H.}\fnms{Hansheng} \snm{Wang}\ead[label=e8]{hansheng@pku.edu.cn}}
    \address{Guanghua School of Management, Peking University\\
    \printead{e8}}
    \thankstext{t2}{Corresponding author.}
\end{aug}

\received{\sday{3} \smonth{9} \syear{2022}}

\begin{abstract}
Labeling mistakes are frequently encountered in real-world applications. If not treated well, the labeling mistakes can deteriorate the classification performances of a model seriously. To address this issue, we propose an improved Na{\"i}ve Bayes method for text classification. It is analytically simple and free of subjective judgements on the correct and incorrect labels. By specifying the generating mechanism of incorrect labels, we optimize the corresponding log-likelihood function iteratively by using an EM algorithm. Our simulation and experiment results show that the improved Na{\"i}ve Bayes method greatly improves the performances of the Na{\"i}ve Bayes method with mislabeled data.
\end{abstract}



\end{frontmatter}

\section{INTRODUCTION}

In the recent decades, artificial intelligence (AI) has fundamentally changed our work and life \citep{AI_impact_2015, AI_impact_2017, AI_impact_2020}.
The success of AI algorithms relies on millions or even billions of labeled data \citep{noisy_datasets_2021}. In the field of automatic face recognition, algorithms can achieve extraordinary performances after being trained on a large-scale labeled dataset \citep{Masi2018Deep}. For example, one popular benchmark dataset for face recognition is \textit{Megaface}, which contains more than 1 million photos. Those photos are from 690,572 individuals \citep{megaface2016}. In the field of autonomous driving, to train an effective self-driving algorithm under different weather and lighting conditions, a large-scale labeled dataset is required. For example, the \textit{nuScenes Dataset} contains 1.4 million images \citep{nuScenes_2020}. Hence, preparing a large-scale labeled dataset is an essential step for real-world AI applications.\\

Currently, the data labeling work mainly depends on human efforts, which is labor-intensive and time-consuming. Estimated by Cognilytica Research \cite{2019Cognilytica}, 25\% of time spent on AI projects is data labeling. To save efforts, many organizations outsource the data labeling work to third-party companies, such as Amazon \citep{MTurk_2011}. In these companies, the labelers should be trained first. However, it is still difficult to guarantee all labelers follow the same labeling criteria. In fact, it is almost impossible to define what is the optimal labeling criteria. For example, \cite{subjective_annotate_1994} find that labelers often apply subjective criteria on medical diagnosis. Furthermore, the high labeling quality often relies on lots of practice. For example, \citep{MSCOCO_2014} find that more experienced labelers often have higher precision and recall rates on category labeling. Therefore, in the early stage of the labeling tasks, labelers are more likely to make labeling mistakes. In some cases, the labeling task needs professional knowledge, such as distinguishing different species of birds \citep{ILSVRC_2015} and annotating medical images \citep{medical_annotate_2021}. Incorrect labels may come from systematic errors of the annotators and inter-annotator variations \citep{CT_2019, medical_annotate_2021}. Consequently, it seems that incorrect labels are inevitable in practice.\\

In fact, incorrect labels have been found among many widely used datasets. For example, \citep{noisy_datasets_2021} reported that 10\% of the labels are incorrect in the \textit{QuickDraw Dataset} and 3.9\% for the \textit{Amazon Reviews Dataset}. Although incorrect labels are common in practice, most existing works have ignored this fact. They directly trained models on the datasets by treating every sample as if it were correctly labeled. Unfortunately, this practice would deteriorate the prediction performance. For example, \citep{ahmad_2019} has trained a multi-layer perceptron classifier on the handwritten Arabic digits dataset with different levels of incorrect labels. Results show that the classification accuracy can be significantly affected by the incorrect labels. Similar empirical evidence can be found for deep neural network models \citep{DLnoise_2017, DNNnoisyrobust_2021}. Therefore, how to account for the mislabeling mechanism in developing classification methods becomes a problem of great importance.\\

There exist plenty of literature that takes the mislabeling issue into consideration. These works can be roughly classified into two categories. They are, respectively, the noise filtering methods and the modified model architectures. The idea of noise filtering methods is to first distinguish between the correctly labeled instances and incorrectly labeled instances by machine learning methods, and then build models only on the correctly labeled instances \cite {filter_1999, SVM_2008, CMTNN_2010, Bayes_noise_detection_2014, MentorNet_2018}. In this regard, how to distinguish the incorrectly labeled instances from the correct ones becomes a problem of critical importance. This leads to two possible solutions. The first one is to treat instances as incorrectly labeled if their predicted labels are different from the observed ones. However, this method could be problematic, because even for correctly labeled instances, their observed labels and predicted labels could still be different. Such a difference is not due to labeling mistakes.
The second solution is to produce an absolutely correctly labeled dataset by human efforts. However, it could be very expensive in human cost and also face inter-labeler variability.\\

Apart from the noise filtering methods, many researchers resort to modifying model architectures to account for the mislabeling issue. BayesANIL \citep{BayesANIL_2005} proposed a Bayesian model for learning with approximate, noisy or incomplete labels.
However, their method required to pre-specify the noise rate, which is hardly known in practice. Other typical examples focus on modifying the deep learning models, such as the Noisy Labels Neural-Network (NLNN) algorithm \citep{DLwithEM_2016}, the decoupling method \citep{decoupling_2017}, and the co-teaching method \citep{coteaching_2018}. However, these modified deep learning models often have a large number of unknown parameters. For example, the NLNN algorithm has more than half a million parameters, and the decoupling method and the co-teaching method both have more than 8 million parameters.\\

To address the above problems, we propose here an improved Na{\"i}ve Bayes model for document classification with incorrect labels. Our algorithm evolves from the Na{\"i}ve Bayes model, which has been verified as arguably one of the most useful tools for document classification \citep{NB_good_2011, Bayes_noise_detection_2014}.
To address the mislabeling issue, we model the mislabeling mechanism by a mislabeling probability matrix. Accordingly, the log-likelihood function can be spelled out. This gives us an opportunity to study the identifiability issue and develop an EM algorithm for model estimation. Simulations and experiments on real-world benchmark datasets show that our algorithm outperforms the traditional Na{\"i}ve Bayes model and some other baseline methods.

\section{RELATED WORK}

As we described before, classification with mislabeled data is a problem of fundamental importance. The existing literature has applied different strategies to tackle this issue. Those methods can be roughly classified into two categories. The first category uses noise filtering methods to produce clean data for the following data analysis tasks. The second category focuses on modifying the architectures of models, so that the mislabeling behavior can be taken into consideration. We summarize related works in each category as follows.\\

\textbf{Noise Filtering Methods.} One way to handle the mislabeled instances is to use noise filtering methods. The basic idea of this type of methods is to train models to distinguish the correctly labeled instances from the mislabeled ones. Then, subsequent data analysis tasks are only accomplished on the correctly labeled instances. For example, \citep{filter_1999} applied a series of learning algorithms, e.g., decision trees, $k$-nearest neighbor classifiers, and linear machines, to construct noise filters. The instances whose predicted labels were different from their observed labels would be discarded as mislabeled data.
\citep{SVM_2008} built a support vector machine as a filter on the training data and removed all instances that are incorrectly classified by the filter. 
\citep{CMTNN_2010} constructed a pair of neural networks called Truth Neural Network (Truth NN) and Falsity Neural Network (False NN) for binary classification problems, thus each label is either 0 or 1. Those two networks had the same structures and inputs. However, the Truth NN was trained on the observed labels. In contrast, the False NN was trained on the complement of the observed labels. The complement of a binary label is defined to be one minus the label. Instances predicted mistakenly by both the Truth NN and the False NN were treated as mislabeled instances.
\citep{Bayes_noise_detection_2014} adopted a Na{\"i}ve Bayes classifier to predict the class label for each training instance based on the highest posterior probability. Then the training instances whose labels are different from the predicted labels were removed from the training dataset to produce noise-free data for the following analysis. \citep{MentorNet_2018} first required a small dataset with all correct labels to learn data-driven curriculums. Then a deep leaning model called MentorNet was trained on this dataset to offer preliminary information on the mislabeled data. Finally, the base deep Convolutional Neural Networks (CNNs) called StudentNet was constructed and updated together with the MentorNet during the training process.\\

The merit of noise filtering methods is that they are straightforward and intuitive. However, they face some challenges.
Specifically, to apply the noise filtering methods, one must distinguish whether an instance is correctly labeled or not.
In this regard, two typical strategies are adopted.
One is to compare the observed label with the predicted label and regard a sample as mislabeled when the two labels do not match.
This strategy is very subjective, because even if all observed labels are correct, most classifiers can still produce wrong predictions.
Thus, when a mismatch occurs, it is difficult to distinguish whether it is a mislabeled sample or just a wrong prediction.
The second strategy is to produce a purely correctly labeled dataset by human efforts, which leads to additional labeling cost.
Again, the human labeling process might also generate incorrect labels. In other words, it is expensive and hard to create a purely correct dataset in practice.\\

\textbf{Modified Model Architecture Methods.}
Another approach to handle the mislabeled instances is to directly modify the model architectures.
The basic idea of this type of methods is to model the mislabeled behaviors under probabilistic or non-parametric frameworks. For example, \citep{BayesANIL_2005} constructed a Bayesian model called BayesANIL. They treated the true labels as latent variables. Then an EM algorithm was applied to deal with the unlabeled and mislabeled instances. \citep{DLwithEM_2016} proposed the Noisy Labels Neural-Network (NLNN) model by assuming the observed labels were created from the true labels by passing through a noisy channel. Subsequently, an extra noise layer was introduced to the neural networks to model the noisy labels. \citep{DLwithEMwithlayer_2017} extended the work of \citep{DLwithEM_2016} by allowing that the appearance of noisy labels depended not only on the true labels, but also on the input features. \citep{decoupling_2017} developed the decoupling method by training two networks. In each mini-batch iteration, the instances with inconsistent predicted labels by the two networks would be picked out.
Then the two networks updated their parameters on the picked out instances. \citep {coteaching_2018} developed the co-teaching method by training two neural networks. Each network fed forward on each mini-batch and selected the data with small losses as the clean data. Then each network back propagated the selected clean data which were offered by its peer network.\\

The merit of these methods is that the incorrectly labeled instances do not need to be distinguished from the correct ones at the model input stage. However, those methods are still accompanied by some issues. First, some models are typically very complicated. For example, the NLNN model contains half a million parameters \citep{DLwithEM_2016, DLwithEMwithlayer_2017}. The decoupling and co-teaching methods are required to update over 8 million parameters \citep{decoupling_2017, coteaching_2018}. Second, there is a strong assumption underlying this strategy. That is, all the incorrect labels should share the same pattern so that a model could be established to discover this pattern and pick out the incorrect labels.
However, in some cases, the incorrect labels happen in a purely random way. Therefore, it is nearly impossible for models to accurately predict those randomly generated incorrect labels.
Finally, some methods require prior knowledge of some important tuning parameters. For example, the application of the BayesANIL method requires the knowledge of the noise rate \citep{BayesANIL_2005}, which is often unknown in practice.\\

\textbf{Our Contributions.}
Inspired by the existing literature, we develop an improved Na{\"i}ve Bayes model for text classification. We focus on text classification because it is a problem of wide applications. The improved Na{\"i}ve Bayes model is analytically simple and free of subjective judgements on the correct and incorrect labels. By assuming the true labels are unobserved and specifying the mechanism of generating incorrect labels, the corresponding log-likelihood functions can be constructed and an EM algorithm can be developed. Simulations and experiments on real-world benchmark datasets show that our algorithm outperforms the traditional Na{\"i}ve Bayes classifier and some other baseline methods.

\section{IMPROVED NA{\"I}VE BAYES WITH MISLABELED DATA}

\subsection{Notations and Model}

Suppose there are a total of $N$ instances. Each instance belongs to one of the $K$ classes. Define $Y_i^* \in \{1, \cdots, K \}$ as the true label of the $i$th instance with $1 \leq i \leq N$. For each instance, we collect a $d$-dimensional binary feature vector. For example, for a document classification problem, each document is viewed as an instance. A feature of a document can be defined as whether the document contains a specific word. If the word occurs in the document, the value of the feature is 1, otherwise 0. For the $i$th instance, the binary feature vector is denoted by $X_i=(X_{i1},...,X_{id})^{\top}$. We denote all the instances by $\mathbb{X} = \{X_i, 1 \leq i \leq N\}$. The true labels for all instances are collected by $\mathbb{Y}^*=\{Y_i^*, 1 \leq i \leq N\}$.\\

To investigate the relationship between the training instances $\mathbb{X}$ and the true labels $\mathbb{Y}^*$, a typical Na{\"i}ve Bayes model can be applied by
maximizing the log-likelihood function $\ln{P(\mathbb{X}, \mathbb{Y}^*)}$. Let $\pi_k=P(Y_i^* = k)$ denote the probability of class $k$. We then have $\sum_{k=1}^K\pi_k=1$. Let $p_{jk}=P(X_{ij}=1|Y_i^* = k)$ denote the probability of the $j$th feature being 1 in class $k$, and $\theta$ be the total parameter set, i.e., $\theta = \{p_{jk}, 1 \leq j \leq d, 1 \leq k \leq K \} \cup \{ \pi_k, 1 \leq k \leq K\}$. To estimate $\theta$, we can construct the log-likelihood function as follows
\begin{align}
    \ell(\theta) = & \ln{P(\mathbb{X}, \mathbb{Y}^*)} = \sum_{i=1}^N \ln{P(X_i, Y_i^*)} \nonumber \\
    =& \sum_{i=1}^N \ln{P(Y_i^*)} + \sum_{i=1}^N \sum_{j=1}^{d} \ln{  P(X_{ij}|Y_i^*)} \nonumber \\
    =& \sum_{i=1}^N   \ln{\pi_{Y_i^*}}  +   \sum_{i=1}^N  \sum_{j=1}^{d} X_{ij}  \ln{p_{j Y_i^*}} \nonumber\\
    &+ \sum_{i=1}^N \sum_{j=1}^{d} (1-X_{ij})  \ln{\left(1-p_{j Y_i^*}\right)} \nonumber.
\end{align}The objective of the Na{\"i}ve Bayes model is to maximize the log-likelihood function $\ell(\theta)$. However, the above estimation is feasible only if all the true labels are observable. Nevertheless, it is common that, the observed labels may be incorrect in reality due to the mislabeling issue. Simply regarding the observed labels as true labels can deteriorate the estimation performances of the Na{\"i}ve Bayes model. We discuss how to address the mislabeling issue in the subsequent sections.

\subsection{Mislabeling Mechanism}

First, we specify the mislabeling mechanism, which facilitates subsequent model estimation. We assume that the true label $Y_i^*$ is a latent variable and the observed label is denoted by $Y_i$. We then denote the observed labels for all instances by $\mathbb{Y}=\{Y_i, 1 \leq i \leq N\}$. Following literature \citep{CCN_2017,CCN_2021}, we assume $P(Y_i|Y_i^*, X_i) = P(Y_i|Y_i^*)$. By doing so, we assume that the mislabeling pattern is random and does not depend on the feature $X_i$ given the true label $Y_i^*$. We next consider how to model the mislabeling mechanism $P(Y_i \neq Y_i^*|Y_i^*)$ appropriately.\\

Define $\rho_{k_1 k_2}$ to be the probability that a labeler labels an instance as class $k_1$ when the true class is $k_2$ with $1 \leq k_1, k_2 \leq K$. This leads to a mislabeling probability matrix as follows
\begin{align}
    \begin{pmatrix}
    \rho_{1 1} & \rho_{1 2} & \cdots &  \rho_{1 K} \\
    \rho_{2 1} & \rho_{2 2} & \cdots & \rho_{2 K} \\
    \vdots & \vdots & \ddots & \vdots \\
    \rho_{K 1} & \rho_{K 2} & \cdots & \rho_{K K} \\
    \end{pmatrix}, \label{conditional matrix}
\end{align}where $\sum_{k_1 = 1}^K \rho_{k_1 k_2} = 1$ for $1 \leq k_2 \leq K$. As one can see, if $\rho_{k k} = 1$ for $1 \leq k \leq K$, then this model reduces to a standard Na{\"i}ve Bayes model. Otherwise, it becomes a new model. If we have $\rho_{k_1 k_2} = \rho_{k_2 k_1}$ for any $k_1 \neq k_2$, we then refer to \eqref{conditional matrix} as a symmetric mislabeling matrix. With this model setup, the whole parameter set becomes $\theta = \{p_{jk}, 1 \leq j \leq d, 1 \leq k \leq K \} \cup \{ \pi_k, 1 \leq k \leq K\} \cup \{ \rho_{k_1 k_2}, 1 \leq k_1, k_2 \leq K \}$. The corresponding log-likelihood function is
{\small
\begin{align}
    \ell(\theta) =& \ln{P(\mathbb{X}, \mathbb{Y}, \mathbb{Y}^*|\theta)} \nonumber \\
    =& \sum_{i = 1}^N \bigg\{ \ln{P(Y_i^*|\theta)} + \ln{P(Y_i|Y_i^*, \theta)} + \sum_{j=1}^d  \ln{P(X_{ij}|Y_i^*, \theta)} \bigg\} \nonumber\\
   =& \sum_{i = 1}^N \ln{\pi_{Y_i^*}} + \sum_{i = 1}^N \ln{\rho_{Y_i Y_i^*}} + \sum_{i = 1}^N \sum_{j=1}^d  X_{ij} \ln{p_{j Y_i^*}} \nonumber\\
   &+ \sum_{i = 1}^N \sum_{j=1}^d (1 - X_{ij}) \ln{(1 - p_{jY_i^*})} \label{ell_theta}.
\end{align}}At the first glance, it seems that we cannot rule out the theoretical possibility that $\rho_{k k} < \max_{k'}\rho_{k' k}$ for $1 \leq k \leq K$ under this mislabeling mechanism. If this happens, there might exist one $k \in \{1, \cdots, K\}/\{Y_i^*\}$, such that $P(Y_i = k|Y_i^*) > P(Y_i = Y_i^*|Y_i^*)$. This suggests that there might exist one mistaken class so that the corresponding mislabeling probability is even larger than that of correct labeling. This seems to be very unreasonable. Our further theoretical analysis reveals that without any constraints on the mislabeling probability matrix in \eqref{conditional matrix}, the mislabeling mechanism might suffer from an identifiability issue. We present the identifiability issue in the following subsection.

\subsection{Identifiability Issue} \label{Identifiability Issue}
As we mentioned before, the mislabeling mechanism might suffer from an identifiability issue. The key reason is as follows. Note that the true labels $\mathbb{Y}^*$ are latent. Without any constrains on the mislabeling probability matrix, any arbitrary
assignment of $\mathbb{Y}^*$ would be acceptable as long as the probability distribution of the observed labels $\mathbb{Y}$ can be replicated. Consider, for example, a classification problem with the true label $Y_i^* \in \{1, \cdots, K\}$ and the observed label $Y_i \in \{1, \cdots, K\}$. Here we shift the true label $Y_i^*$ to $Y_i^* + 1$ for $Y_i^* \in \{ 1, \cdots, K-1\}$ and replace $Y_i^* = K$ with 1. We denote the shifted labels by $\widetilde{Y}_i^*$ and collect all the shifted labels into $\widetilde{\mathbb{Y}}^*$. Denote the one-to-one mapping from $Y_i^*$ to $\widetilde{Y}_i^*$ by $g(\cdot)$. Recall that the log-likelihood function based on $\mathbb{Y}^*$ is given in \eqref{ell_theta}. However, with a shift of $\rho_{k_1 k_2}$s and $p_{jk}$s, the value of the new log-likelihood function based on $\widetilde{\mathbb{Y}}^*$ remains the same. The new labeling probability $\widetilde{\rho}_{k_1 k_2}$ satisfies $\widetilde{\rho}_{k_1 k_2} = {\rho}_{k_1 g^{-1}(k_2)}$ and the new feature probability $\widetilde{p}_{jk}$ satisfies $\widetilde{p}_{jk} = {p}_{j g^{-1}(k)}$. Denote the new parameter set as $\widetilde{\theta}$. Consequently, the new log-likelihood function based on $\widetilde{\mathbb{Y}}^*$ becomes
{\small
\begin{align}
    \widetilde{\ell}\Big(\widetilde{\theta}\Big) =& \ln{\Big(\mathbb{X}, \mathbb{Y}, \widetilde{\mathbb{Y}}^* \Big|\widetilde{\theta} \Big)} \nonumber \\
    =& \sum_{i=1}^N \! \bigg\{ \! \ln \! P\Big(\widetilde{Y}_i^*\Big|\widetilde{\theta}\Big) \! + \! \ln \! P\Big(Y_i\Big|\widetilde{Y}_i^*, \widetilde{\theta}\Big) \! + \! \sum_{j=1}^d \ln \! P\Big(X_{ij} \Big| \widetilde{Y}_i^*, \widetilde{\theta}\Big)\! \bigg\} \nonumber \\
    =& \sum_{i=1}^N \ln{\pi_{\widetilde{Y}_i^*}} + \sum_{i = 1}^N \ln{ \widetilde{\rho}_{Y_i \widetilde{Y}_i^*}} + \sum_{i=1}^N \sum_{j=1}^d X_{ij} \ln{\widetilde{p}_{j \widetilde{Y}_i^*}} \nonumber \\
    &+ \sum_{i = 1}^N \sum_{j=1}^d (1 - X_{ij}) \ln{(1 - \widetilde{p}_{j \widetilde{Y}_i^*})}.\nonumber
\end{align}}One can easily verify that the resulting log-likelihood values are exactly the same as before, i.e., $\ell(\theta) = \widetilde{\ell}(\widetilde{\theta})$. This suggests that the observed probabilistic behavior about $\mathbb{Y}$ and $\mathbb{X}$ can be equally well explained by either $\mathbb{Y}^*$ or $\widetilde{\mathbb{Y}}^*$ with different 
parameters. As a consequence, the model under this mislabeling mechanism might not be uniquely identified. Then how to solve this identifiability issue becomes an important problem. In this regard, we notice that a qualified labeler should not perform too badly, in the sense that the diagonal probabilities should be larger than the off-diagonal probabilities. Otherwise, there might exist one class $k_1 \neq k_2$ satisfying $P(Y_i = k_1 | Y_i^* = k_2) > P(Y_i = k_2 | Y_i^* = k_2)$. Thus, we are motivated to assume that the diagonal element $\rho_{k k}$ is larger than the off-diagonal element $\rho_{k' k}$ with $k' \neq k$, i.e., $\rho_{kk} > \max_{k' \neq k} \rho_{k'k}$ for every $1 \leq k \leq K$. This makes the model based on \eqref{conditional matrix} identifiable.

\subsection{Mislabeling Impact}\label{Mislabeling Impact}

As we demonstrated before, the incorrect labels might/might not hurt the performances of the models severely \citep{ahmad_2019}. In this subsection, we study the mislabeling impact theoretically and show when the mislabeling has little impact on the predictions and when it is a big issue. We start with a warm-up case with $K=2$.\\

\textbf{Two-Class Case}. When there are only two classes, the mislabeling probability matrix in (\ref{conditional matrix}) becomes a $2 \times 2$ matrix. For simplicity, we consider a balanced sample size in each class here with $P(Y^*_i = 1) = P(Y^*_i = 2) = 0.5$. For illustration purpose, we further assume that $\rho_{12} = \rho_{21}$ so that $P(Y_i = 1) = P(Y_i=2) = 0.5$. Recall that the probability of the $j$th feature being 1 in class $k$ is denoted by $p_{jk} \triangleq P(X_{ij}=1|Y^*_i=k)$ for any $k \in \{1, 2\}$. 
For a more intuitive understanding, assume without loss of generality that $p_{j1} > p_{j2}$. This suggests that the random event $X_{ij} = 1$ is more likely to happen with $Y_i^* = 1$ than $Y_i^* = 2$. This further implies that $P(Y_i^*=1|X_{ij}=1) > P(Y_i^*=2|X_{ij}=1)$ if equal prior probability can be assumed for $Y_i^*$. Consequently, if $X_{ij}=1$ is the only information we have, we should naturally predict $Y_i^*$ to be 1. Then, it is of great interest to query how this might be affected by mislabeling. We are particularly interested in knowing the relative order of $P(Y_i = 1|X_{ij} = 1)$ and $P(Y_i = 2|X_{ij}=1)$. We then have
{\small
\begin{align}
    &P(Y_i = 1|X_{ij}=1) - P(Y_i = 2|X_{ij}=1) \nonumber \\
    =& \ 0.5 \times (p_{j1} - p_{j2}) \times (\rho_{11} - \rho_{12}) / P(X_{ij}=1) > 0, \label{Y=1 - Y=2}
\end{align}}since $p_{j1} - p_{j2} > 0$ and $\rho_{11} - \rho_{12} > 0$ by model assumption. By \eqref{Y=1 - Y=2}, we find that the relative order of $P(Y_i=1|X_{ij}=1)$ and $P(Y_i=2|X_{ij}=1)$ is not changed by mislabeling. Consequently, if $X_{ij}=1$ is the only information we have, we should naturally predict $Y_i$ to be 1. This prediction result is the same as that of $Y_i^*$. This suggests that the adverse effect caused by the mislabeling seems to be very limited for this particular special case.\\

\textbf{Multiple-Class Case with Constant Mislabeling Probability}. The above discussion suggests that the adverse effect due to mislabeling could be very small if $K=2$ and if the mislabeling probability matrix is symmetric. We are then inspired to study the case with $K \geq 3$. For simplicity purpose, we consider a case with a constant mislabeling probability. That is to assume that $\rho_{kk} = \rho$ for some $\rho \in (0.5, 1)$ and $\rho_{k_1 k_2} = (1-\rho) / (K-1)$ for any $k_1 \neq k_2$. We consider here equal prior probability for each class. 
Without loss of generality, we can assume that $p_{j k_1} > p_{j k_2}$. This implies that the random event $X_{ij} = 1$ is more likely to happen with $Y_i^* = k_1$ than $Y_i^* = k_2$. Then it is more reasonable to predict $Y_i^*=k_1$ than $Y_i^*=k_2$ if $X_{ij}=1$ is observed. In fact, it can be verified that $P(Y_i^* = k_1|X_{ij}=1) > P(Y_i^* = k_2|X_{ij}=1)$ if equal prior probability for $Y_i^*$ can be assumed. Specifically, we can also compute $P(Y_i=k_1|X_{ij} > 1) - P(Y_i=k_2|X_{ij} > 1)$ by
{\small
\begin{align}
    &P(Y_i=k_1|X_{ij} > 1) - P(Y_i=k_2|X_{ij} > 1) \nonumber \\
    =& \ \left(\frac{1}{K}\right) \times (p_{j k_1} - p_{j k_2}) \times \left(\frac{K\rho-1}{K-1}\right) / P(X_{ij}=1) > 0, \label{Yk1 - Yk2}
\end{align}}with $K \geq 3$ and $\rho > 0.5$. By \eqref{Yk1 - Yk2}, the relative order of $P(Y_i=k_1|X_{ij}=1)$ and $P(Y_i=k_2|X_{ij}=1)$ remains the same as that of $P(Y_i^*=k_1|X_{ij}=1)$ and $P(Y_i^*=k_2|X_{ij}=1)$ for any two classes $k_1 \neq k_2$. Thus, the predictions of the Na{\"i}ve Bayes model based on the observed labels are the same as that of the model based on the true labels. This suggests that the adverse effect due to mislabeling should be very limited as long as the mislabeling probability is constant.
\\

\textbf{Multiple-Class Case with Varying Mislabeling Probability}. We next study the case with $K \geq 3$ and non-constant mislabeling probabilities. Specifically, we consider here a special $K \geq 3$ case. The mislabeling probability matrix is given as follows
\begin{align}
    \begin{pmatrix}
    \rho & 1-\rho & 1-\rho & \cdots & 1-\rho \\
    1-\rho & \rho & 0 & \cdots & 0  \\
    0 & 0 & \rho & \cdots & 0 \\
    \vdots & \vdots & \vdots & \ddots & \vdots \\
    0 & 0 & 0 & \cdots & \rho \\
    \end{pmatrix}, \label{new labeling matrix}
\end{align}where $0.9 \leq \rho < 1$. In this case, we require that $K \gg \rho / (1 - \rho) + 1 \geq 11$. By this mislabeling probability matrix, we assume that class 1 is the most confusing one.
Specifically, we know that $\rho_{k_1 k_2} = 0$ for any $k_1 > 2$, $k_2>2$, and $k_1 \neq k_2$. In other words, no mislabeling happens between any class pairs, unless one of the class is class 1. If the true class $Y_i^* = 1$, then with probability $\rho$ the observed class $Y_i$ remains to be 1 and with probability $(1-\rho)$ the observed class $Y_i$ becomes 2. If the true class $Y_i^*$ is class $k \geq 2$, then with probability $\rho$ the observed class $Y_i$ remains to be $k$ and with probability $(1-\rho)$ the observed class $Y_i$ becomes 1.
For simplicity, we assume $p_{j1} \ll p_{jk}$ for $k \geq 2$. This leads to $P(Y_i^*=1|X_{ij}=1) < P(Y_i^*=k|X_{ij}=1)$. 
However, one can easily verify that 
{\small
\begin{align*}
    & P(Y_i=1|X_{ij}=1) - P(Y_i=k|X_{ij}=1) \nonumber \\
    =& \left(\frac{1-\rho}{K}\right) \times \bigg\{ \left(\frac{\rho}{1 - \rho}\right) p_{j1} + (K-1)p_{j2} - \left(\frac{\rho}{1 - \rho}\right) p_{j2}\bigg\} > 0,
\end{align*}}for any $k>2$. Thus, serious adverse effect is caused by the mislabeling mechanism. Consequently, the Na{\"i}ve Bayes model can hardly differentiate class 1 from class $k > 2$. This special case suggests that when incorrectness accumulates on a few number of classes, the Na{\"i}ve Bayes model might have trouble in correctly identifying those classes. This is the case where a small fraction of incorrect labels can still cause big trouble and the problem cannot be solved by simply enlarging sample sizes.\\

Note that the labeling mistakes may severely deteriorate the classification accuracy. Then how to evaluate the mislabeling effect becomes a problem of great importance. Note that the classification accuracy is an important metric for classification performance evaluation. Denote the classification accuracy based on $(\mathbb{X}, \mathbb{Y})$ by $\text{ACC}$. Denote the classification accuracy based on $(\mathbb{X}, \mathbb{Y}^*)$ by $\text{ACC}^*$. We can define their difference $\Delta \text{ACC} =\text{ACC} - \text{ACC}^*$ as a simple measure for mislabeling effect.

\subsection{An EM Algorithm} \label{EM algorithm}

As we demonstrated in the {Subsection \ref{Identifiability Issue}}, the log-likelihood function derived by the mislabeling mechanism is identifiable by assuming that the mislabeling probability matrix in \eqref{conditional matrix} satisfies $\rho_{kk} > \max_{k' \neq k} \rho_{k'k}$ for every $1 \leq k \leq K$. Specifically, we consider how to estimate the corresponding model by optimizing the log-likelihood function $\ell(\theta)$. Since the true labels $\mathbb{Y}^*$ are unknown, $\ell(\theta)$ cannot be optimized directly. Therefore, we develop an EM algorithm to iteratively estimate the parameters and optimize the log-likelihood function. In each iteration, an expectation step (E step) and a maximization step (M step) are conducted. Below, we discuss each step in details.\\

\textbf{E step}. In the $t$th iteration, we compute the expectation of the log-likelihood $\ell(\theta)$ given the observed data $(\mathbb{X}, \mathbb{Y})$ and the current parameter estimate $\widehat{\theta}^{(t)}$. The conditional expectation can be derived as
{\small
\begin{align}
    Q\Big(\theta, \widehat{\theta}^{(t)}\Big) =& E\Big\{\ell(\theta)\Big|\mathbb{X}, \mathbb{Y},\widehat{\theta}^{(t)}\Big\} \nonumber \\
    =& \sum_{i=1}^N \sum_{k=1}^K \widehat{\gamma}_{ik}^{(t)} \ln{\pi_k} + \sum_{i=1}^N \sum_{k=1}^K \widehat{\gamma}_{i k}^{(t)} \ln{\rho_{Y_i k}} \nonumber \\
    &+ \sum_{i=1}^N \sum_{j=1}^d \sum_{k=1}^K X_{ij} \widehat{\gamma}_{ik}^{(t)} \ln{p_{jk}} \nonumber \\
    &+ \sum_{i=1}^N \sum_{j=1}^d \sum_{k=1}^K (1-X_{ij})\widehat{\gamma}_{ik}^{(t)}  \ln{(1-p_{jk})}, \label{Q}
\end{align}}where $\widehat{\gamma}_{ik}^{(t)}$ denotes $ P(Y_i^*=k|X_i, Y_i, \widehat{\theta}^{(t)})$. The computation of (\ref{Q}) depends on $\widehat{\gamma}_{ik}^{(t)}$ for $1 \leq i \leq N$ and $1 \leq k \leq K$. To compute $\widehat{\gamma}_{ik}^{(t)}$, we first define $\widehat{\zeta}_{ik}^{(t)}$ as follows
{\small
\begin{align}
    \widehat{\zeta}_{ik}^{(t)} = \widehat{\pi}_k^{(t)} \widehat{\rho}_{Y_i k}^{(t)}
    \prod_{j=1}^d \widehat{p}_{jk}^{(t) X_{ij}} \Big\{1 - \widehat{p}_{jk}^{(t)}\Big\}^{1 - X_{ij}}.
\end{align}}Then we compute $\widehat{\gamma}_{ik}^{(t)}=\widehat{\zeta}_{ik}^{(t)}/\sum_{k=1}^K \widehat{\zeta}_{ik}^{(t)}$.\\

\textbf{M step}. In this step, we maximize the conditional expectation $ {Q(\theta, \widehat{\theta}^{(t)})}$ to get the new estimate $\widehat{\theta}^{(t+1)}$. Note that $\sum_{k=1}^K \pi_k = 1$ and $\sum_{k_1=1}^K \rho_{k_1 k_2} = 1$ for $1 \leq k_2 \leq K$. Based on the Lagrange multiplier method, we can define the Lagrangian function as
{\small 
\begin{align*}
    Q = Q\left(\theta, \widehat{\theta}^{(t)}\right) + \lambda \left( 1 - \sum_{k=1}^K \pi_k \right) + \sum_{k_2=1}^K \lambda_{k_2} \left(1 -  \sum_{k_1 = 1} \rho_{k_1 k_2}\right),
\end{align*}}where $\lambda$ and $\lambda_{k_2}$ with $1 \leq k_2 \leq K$ are additional parameters used in the Lagrange multiplier method. After maximizing the Lagrangian function $Q$, the updated parameters in the $(t+1)$th iteration are computed as
{\small
\begin{align*}
    & \widehat{\pi}_k^{(t+1)} = \sum_{i=1}^N \widehat{\gamma}_{ik}^{(t)} / N , \ 1 \leq k \leq K, \nonumber \\
    & \widehat{p}_{jk}^{(t+1)} = \left(\sum_{i=1}^N X_{ij} \widehat{\gamma}_{ik}^{(t)}\right) \bigg/ \left(\sum_{i=1}^N \widehat{\gamma}_{ik}^{(t)}\right), \ 1 \leq j \leq d,
    \ 1 \leq k \leq K, \nonumber \\
    & \widehat{\rho}^{(t+1)}_{k_1 k_2} = \left(\sum_{i=1}^N I(Y_i = k_1) \widehat{\gamma}_{i k_2}^{(t)}\right) \bigg/ \left(\sum_{i=1}^N \widehat{\gamma}_{i k_2}^{(t)}\right), \  1 \leq k_1, k_2 \leq K.
\end{align*}To initialize the parameters, we could set $\widehat{\pi}_k = 1/K$ for $1 \leq k \leq K$. In addition, for $\widehat{p}_{jk}$ ($1 \leq j \leq d, 1 \leq k \leq K$) we use random initialization. We conduct random initialization for $\widehat{\rho}_{k_1 k_2}$ with the diagonal elements larger than 0.5 for $1 \leq k_1, k_2 \leq K$. Subsequently, the E step and the M step are conducted iteratively until convergence or the pre-specified maximum iteration steps are reached.
}

\section{EXPERIMENTS}

\subsection{Simulation Experiments \label{simulation}}

Numerical studies are conducted to evaluate the finite sample performances of the proposed method.\\

\textbf{Simulation Data}. We generate synthetic data and set the dimension of the instances to be $d=500$. Each instance belongs to one of the $K=5$ classes. The data size is set to be $n =500$, 1000, and 5000. 20\% of the simulation data are split into the testing set. The prior probability $\pi_k$ is set to be $1/K$ for any $1 \leq k \leq K$. An unbalanced case is included in Appendix \ref{unbalanced sample size}. The diagonal elements of the mislabeling probability matrix are randomly generated from different intervals, which are shown in Table \ref{simulation-table}.
The off-diagonal elements are then randomly sampled from $[0, 1)$ such that $\sum_{k_1=1}^K \rho_{k_1 k_2} = 1$ for any $1 \leq k_2 \leq K$. The $p_{jk}$s are generated by adding up uniform random numbers from $[0, 0.1)$ and normal random numbers with mean 0.65 and standard deviation 0.06. In each simulation setting, the experiment is repeated for $B=100$ times. 
\\

\textbf{Baseline Methods}. We compare the improved Na{\"i}ve Bayes method (INB) with: 1) the Na{\"i}ve Bayes (NB) model without considering the incorrect labeling; 2) NLNN method of \cite{DLwithEM_2016}; 3) NAL\footnote{{https://github.com/udibr/noisy\_labels}} method of \cite{DLwithEMwithlayer_2017}; 4) the NB model using the true values of the parameters (NB-T).\\

\textbf{Evaluation Metrics}. We evaluate the proposed method from two perspectives. First, we use the mean squared error (MSE) to evaluate parameter estimation accuracy. The MSE is defined mean square of the difference between estimated parameters and the true values of the parameters. $\text{MSE} = \sum_{j=1}^d \sum_{k=1}^K (\widehat{p}_{jk} - p_{jk})^2 /(dK)$, which measures the estimation accuracy.
Second, we use the classification accuracy (ACC) on the testing set to evaluate the classification accuracy, i.e., $\text{ACC} = s / n_0 \times 100 \%$, where $s$ is the number of correctly predicted labels in the testing set and $n_0$ is the size of the testing set. 
Third, we also compute the AUC\footnote{\href{}{https://scikit-learn.org/stable/modules/model\_evaluation.html\#roc-metrics}} (i.e., the area under the receiver operating characteristic curve, ROC) of the methods \citep{microAUC_2017}. The AUC we used here is the macro-AUC metric, which is a multi-class performance measure.  Fourth, We also evaluated the mislabeling effect $\Delta$ACC on the simulation data, which is defined in Section \ref{Mislabeling Impact}. 
\\

\begin{table*}
\caption{Finite sample performances of different methods with different $\rho_{kk}$ intervals and sample sizes.}\label{simulation-table}
{\small
\begin{tabular}{@{}c|c|cc|ccccc|ccccc|c@{}}
\toprule
                                &      & \multicolumn{2}{c|}{MSE$\times 10^{-3}$} & \multicolumn{5}{c|}{ACC (\%)}    & \multicolumn{5}{c|}{AUC (\%)}    &                  \\
$\rho_{kk}$ Intervals           & $n$  & NB                  & INB                & NB   & INB  & NB-T & NLNN & NAL  & NB   & INB  & NB-T & NLNN & NAL  & $\Delta$ACC (\%) \\ \midrule
\multirow{3}{*}{{[}0.55, 0.65)} & 500  & 3.3                 & 2.9                & 66.0 & 83.2 & 96.6 & 20.7 & 28.3 & 90.7 & 97.2 & 99.7 & 52.7 & 63.7 & -22.3            \\
                                & 1000 & 2.2                 & 1.2                & 75.9 & 92.6 & 96.9 & 21.1 & 39.3 & 94.7 & 99.4 & 99.8 & 55.4 & 74.8 & -17.6            \\
                                & 5000 & 1.0                 & 0.2                & 90.9 & 95.0 & 95.7 & 30.8 & 80.6 & 99.1 & 99.7 & 99.8 & 64.4 & 96.5 & -4.2             \\ \midrule
\multirow{3}{*}{{[}0.65, 0.75)} & 500  & 3.0                 & 2.8                & 73.8 & 84.0 & 96.6 & 20.7 & 30.5 & 94.2 & 97.7 & 99.7 & 53.8 & 67.8 & -14.5            \\
                                & 1000 & 1.7                 & 1.2                & 85.3 & 92.7 & 96.9 & 21.1 & 46.8 & 97.9 & 99.4 & 99.8 & 56.4 & 80.6 & -8.1             \\
                                & 5000 & 0.7                 & 0.2                & 93.3 & 95.1 & 95.7 & 35.3 & 86.1 & 99.5 & 99.7 & 99.8 & 66.7 & 98.0 & -1.8             \\ \midrule
\multirow{3}{*}{{[}0.75, 0.85)} & 500  & 2.7                 & 2.7                & 78.7 & 85.6 & 96.6 & 20.8 & 32.9 & 96.0 & 98.1 & 99.7 & 55.0 & 70.0 & -9.6             \\
                                & 1000 & 1.4                 & 1.2                & 89.2 & 93.0 & 96.9 & 22.0 & 54.6 & 98.8 & 99.4 & 99.8 & 58.6 & 85.2 & -4.2             \\
                                & 5000 & 0.4                 & 0.2                & 94.2 & 95.1 & 95.7 & 34.9 & 88.6 & 99.6 & 99.7 & 99.8 & 65.6 & 98.6 & -0.9             \\ \midrule
\multirow{3}{*}{{[}0.85, 0.95)} & 500  & 2.4                 & 2.5                & 84.3 & 86.8 & 96.6 & 21.7 & 36.8 & 97.9 & 98.4 & 99.7 & 56.7 & 73.3 & -4.0             \\
                                & 1000 & 1.2                 & 1.2                & 91.8 & 93.2 & 96.9 & 22.3 & 60.0 & 99.3 & 99.5 & 99.8 & 58.7 & 88.1 & -1.6             \\
                                & 5000 & 0.3                 & 0.2                & 94.7 & 95.1 & 95.7 & 35.6 & 90.1 & 99.7 & 99.7 & 99.8 & 65.2 & 99.0 & -0.4             \\ \midrule
\multirow{3}{*}{{[}1.0, 1.0{]}} & 500  & 2.3                 & 2.4                & 88.2 & 88.0 & 96.6 & 22.1 & 40.0 & 98.8 & 98.7 & 99.7 & 56.9 & 76.2 & 0.0              \\
                                & 1000 & 1.1                 & 1.1                & 93.4 & 93.3 & 96.9 & 24.0 & 64.5 & 99.5 & 99.5 & 99.8 & 61.7 & 90.4 & 0.0              \\
                                & 5000 & 0.2                 & 0.2                & 95.1 & 95.1 & 95.7 & 31.2 & 92.1 & 99.7 & 99.7 & 99.8 & 62.4 & 99.3 & 0.0              \\ \bottomrule
\end{tabular}
}
\end{table*}

\begin{figure}[ht]
\centering
\includegraphics[scale=0.6]{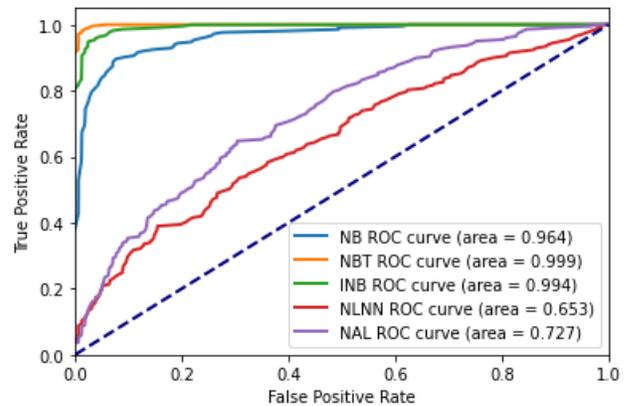}
\vskip -0.1in
\caption{An arbitrarily selected illustrative example of the ROC curve with $n = 1000$ and $\rho_{kk} \in [0.55, 0.65)$.  }\label{fig:roc}
\vskip -0.2in
\end{figure}

\textbf{Simulation Results with $0.5 < \rho_{kk} < 1$}. 
From the last column of $\Delta$ACC in Table \ref{simulation-table}, we know that as $\rho_{kk}$s increase, the mislabeling effect $\Delta$ACC decreases. It is reasonable because with the increase of $\rho_{kk}$s, the number of the incorrect labels decreases, so that the mislabeling effect is relieved.
The MSE, ACC, and AUC results under different $\rho_{kk}$ intervals and data sizes are presented in Table \ref{simulation-table}. An example of the ROC curve \citep{roc_2006} is plotted in Figure \ref{fig:roc}, which is an average of the ROC curves for different classes.
First, we observe that with the increase of data size, the performances of the different methods become better. Next, when $\rho_{kk}$s increase, the average performances of different methods become better and closer.
Since the true values of the parameters are unknown in practice, the NB-T method is not feasible in real-world applications. In the simulation study, the performances of the NB-T method can act as an important reference for the theoretically optimal classification performances. The performances of the INB method are closer to that of the NB-T method than the NB method.
Last, besides the NB-T method, the INB method outperforms the other methods across all simulation settings most of the time. Specifically, under different $\rho_{kk}$ intervals and data sizes, the INB method has achieved smallest MSE values, highest ACC values, and highest AUC values on the testing sets most of the time. This illustrates the advantage of the proposed INB method.\\

\textbf{Simulation Results with $\rho_{kk} = 1$}.
Note that if $\rho_{kk}=1$ for any $1 \leq k \leq K$, there are no incorrect labels generated in the simulation experiments. Thus, the mislabeling effect $\Delta$ACC is naturally 0. When no mislabeling happens, the performances of the NB, INB, and NB-T become identical and optimal.

\subsection{Real Data Experiments}

\begin{figure}
\centering
\vskip -0.2in
\includegraphics[scale=0.6]{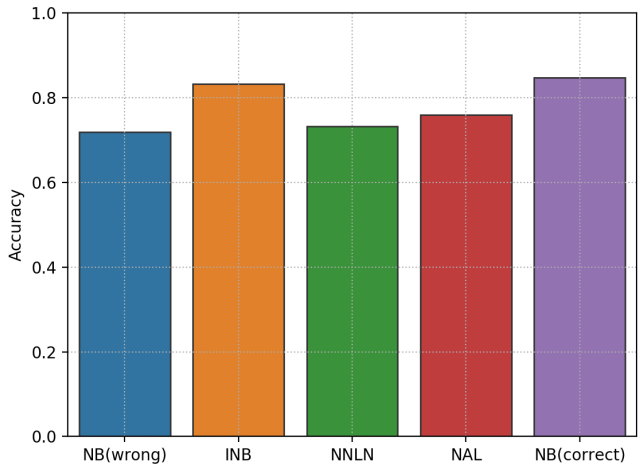}
\caption{Classification accuracy results on the \textit{20 Newsgroups Dataset}. }\label{fig:20news}
\vskip -0.2in
\end{figure}

\textbf{The 20 Newsgroups Benchmark Dataset}. We compare the performances of each method on an important benchmark dataset \textit{20 Newsgroups} \citep{20News_1995}. It is a dataset containing 18,846 documents with 15,076 in the training set and 3,770 in the testing set. First, we construct a dictionary of 130,107 words for the documents. Next, we compute the TF-IDF value for each word and maintain the top 7,302 words with the highest TF-IDF values. To gain some intuitive understanding about the features (or words), we list the top 10 representative features in Table \ref{table:20news features}.
The mislabeling samples have been artificially created by \cite{wang_2021} and the noise rates range from 0.1 to 0.5. We follow their practice to set mislabeling rate as 20\%. The detailed results are given in Figure \ref{fig:20news}. We compare the performances of the five methods: 1) NB(wrong), a NB model trained on the training set, which contains wrong labels; 2) INB method; 3) NLNN method; 4) NAL method; 5) NB(correct), a NB model trained on the training data with correct labels. The mislabeling effect $\Delta\text{ACC}$ on the \textit{20 Newsgroups} dataset with 20\% mislabeling rate is -12.85\%. We find that the performances of the INB method are very close to NB based on the correct labels, outperforming the other methods.\\

\begin{table*}[ht]
\caption{Top 10 representative features of the \textit{20 Newsgroups} datset.}\label{table:20news features}
\begin{tabular}{@{}ccccccccccc@{}}
\toprule
Doc ID & windows & NASA & god & drive & apple & IBM & car & Virginia & MIT & space \\ \midrule
5554   & 1       & 1    & 1   & 1     & 1     & 1   & 0   & 1        & 1   & 1     \\
1117   & 1       & 1    & 1   & 1     & 1     & 1   & 0   & 1        & 1   & 1     \\
7404   & 1       & 1    & 0   & 1     & 1     & 1   & 0   & 1        & 1   & 1     \\
2086   & 1       & 1    & 0   & 0     & 0     & 1   & 0   & 0        & 1   & 0     \\
4770   & 1       & 1    & 0   & 0     & 0     & 1   & 0   & 0        & 1   & 1     \\ \bottomrule
\end{tabular}
\end{table*}

\textbf{Live Streaming Dialog Dataset}\footnote{\href{https://github.com/Helenology/Improved-Naive-Bayes-with-Mislabeld-Data/blob/main/data0126.csv}{https://github.com/Helenology/Improved-Naive-Bayes-with-Mislabeld-Data/blob/main/data0126.csv}}. To further illustrate our method, we present here another interesting real dataset analysis generously donated by our industry collaborators. During the epidemic of COVID-19, a lot of consumers turn to live streaming platforms such as TikTok to watch the streamers introducing products. Consumers are free to ask questions about the products by sending live comments or messages to the streamers. Responding to those comments or messages immediately and appropriately is likely to increase the product sales. However, when considerable consumers are sending live comments or messages in a very short time, it is a great burden for the streamers to respond immediately. Such a burden could be relieved by building an auto-responder bot. One possible workflow for an efficient auto-responder bot is that it first classifies a message into a specific category and then responds according to the category.\\

\begin{table*}
\caption{Thirteen Categories that the messages are classified into and their corresponding responses. }\label{category}
{ 
\begin{tabular}{@{}cll@{}}
\toprule
Category Number & Category Description                          & Response Strategy                                     \\ \midrule
1               & Questions related to loans                    & Ask for the consumer's contact information.               \\
2               & Questions related to discounts                & Directly reply ``The discount is XX\%".               \\
3               & Questions related to car prices               & Directly reply ``The car price is XX RMB".            \\
4               & Questions related to total cost               & Directly reply ``The total cost is XX RMB".           \\
5               & Questions related to availability             & Directly reply ``The car is available/unavailable".   \\
6               & Questions related to license plate            & Answer ``Yes" for the same province/``No" otherwise.  \\
7               & Questions related to store address            & Directly reply the store address.                     \\
8               & Questions totally irrelevant                  & Ignore the message and do not reply.                  \\
9               & Leaving contact information                   & Directly reply ``Message received".                   \\
10              & Asking for contact information                & A salesman/saleswoman will be automatically assigned. \\
11              & Greeting message without car information      & Ask for the consumer's car preference.                    \\
12              & Messages without configuration information    & Ask for the consumer's configuration preference.          \\
13              & Unclear message about new or second-hand cars & Directly ask ``Do you mean a new or second-hand car". \\ \bottomrule
\end{tabular}
}
\end{table*}

The objective of this study is to design a system, which answers consumers' questions automatically. To this end, we need to classify those messages into different categories according to a carefully designed reply strategy; see the \engordnumber{3} column in Table \ref{category}. For example, all the questions related to loans are formed into one category. Whenever a message is correctly classified into this category, the automatic system should reply to the consumer according to standard procedures, which are listed in Figure \ref{fig:flowchart} in Appendix \ref{flowchart}.
The automatic reply system is a sophisticated system, which has experienced a huge amount of tests and improvements. This leads to a total of 13 categories together with their automatic replying strategy; see Table \ref{category} for the details.\\

Subsequently, volunteers are used to interact with a preliminarily designed auto-responder bot and collect abundant messages, most of which are relevant to purchasing cars of one particular brand. Then, volunteers are required to label the messages according to the pre-defined 13 categories. However, we found that the labels are not always correct. For research purpose, all those labels are manually and carefully checked by field experts. This leads to a set of carefully checked labels, which could be considered as the golden standard. This leads to a dataset with $N=1416$ messages. For each message, both the observed label $Y_i$ and the true label $Y_i^*$ are provided. After a careful inspection, we find that about 19.49\% of the observed labels are incorrect. \\

For modeling the classification problem, we extract a feature vector of $d = 22$ dimension from the text information of the message.
Each feature is a binary feature carefully defined by the field experts, with the pre-defined 13-category-based auto response strategy taken into consideration; see the last column in Table \ref{category}. This is a procedure containing multiple important steps. First, several field experts from a live streaming platform provide us with basic keyword dictionaries for different categories. Next, the basic keyword dictionaries are used for preliminary classifications. Last, the missed yet important keywords from the wrong predictions are added into the keyword dictionaries. This leads to the final set of keywords used for feature definitions. For example, many consumers are interested in discount information. As a result, the corresponding messages should contain keywords like ``discount" or ``cheaper". Then, a binary feature can be defined as $X_{17}$ to indicate whether the consumer is interested in discount information or not. As one can see, this is a feature not only useful for the subsequent classification task but also can be nicely interpreted. For a more intuitive understanding, the detailed descriptions for all those features of the \textit{Live Streaming Dialog Dataset} are listed in Table \ref{independent table}. There are a total of 22 independent binary variables. If a message satisfies the conditions listed in the \engordnumber{2} column in Table \ref{independent table}, then the corresponding variable is 1, otherwise 0.
\\

\begin{table*}[ht]
\caption{Descriptions for the independent variables. }\label{independent table}
{
\begin{tabular}{@{}cc@{}}
\toprule
Variable Name             & The Practical Meaning                                                            \\ \midrule
$X_1$                     & Whether the message contains car information only?                               \\
$X_2$                     & Is the message a question?                                                       \\
$X_3$                     & Is the message the first message sent by the consumer?                           \\
$X_4$                     & Whether this message is about one specific car?                                  \\
$X_5$                     & Whether detailed car configuration information is provided in the message?       \\
$X_6$                     & Is configuration information included in the message?                            \\
$X_7$                     & Is this a message about the car store address?                                   \\
$X_8$                     & Whether the consumer's contact information is given in the message?              \\
$X_9$                     & Whether this message is about a new car?                                         \\
$X_{10}$                  & Whether the message is about a second-hand car?                                  \\
$X_{11}$                  & Does the consumer ask for contact information in this message?                   \\
$X_{12}$                  & Is the message a statement about one specific car?                               \\
$X_{13}$                  & Has the consumer left his contact information in the previous messages?          \\
$X_{14}$                  & Is the message a question on license plates?                                     \\
$X_{15}$                  & Is the message a question on total cost?                                         \\
$X_{16}$                  & Is the message a question on car prices?                                         \\
$X_{17}$                  & Is the message a question on discounts?                                          \\
$X_{18}$                  & Is the message a question on whether the car is available or needs reservations? \\
$X_{19}$                  & Is the message a question on loans?                                              \\
\multirow{3}{*}{$X_{20}$} & Is the message not about the loan, total cost, car price, discount, asking for   \\
                          & contact information, leaving contact information, availability,                  \\
                          & car store address, or license plate?                                             \\
\multirow{4}{*}{$X_{21}$} & Is the message not about the loan, total cost, car price, discount, asking for   \\
                          & contact information, leaving contact information, availability,                  \\
                          & car store address, or license plate?                                             \\
                          & Is the message the first message sent by the consumer?                           \\
\multirow{5}{*}{$X_{22}$} & Is the message not about the loan, total cost, car price, discount, asking for   \\
                          & contact information, leaving contact information, availability,                  \\
                          & car store address, or license plate?                                             \\
                          & Is the message the first message sent by the consumer?                           \\
                          & Can we tell which car the consumer refers to in this message?                    \\ \bottomrule
\end{tabular}
}
\end{table*}

\textbf{Experimental Setup}.
We randomly assign 80\% of the data to the training set, while the rest 20\% are assigned to the testing set. The training set contains both correct and incorrect labels. However, the testing set only contains the correctly labeled data to evaluate the classification accuracy. The split of the training set and testing set is repeated for $B = 100$ times. Under each split, we compare the classification accuracy of five methods: 1) NB(wrong), which is a NB model trained on the training set containing incorrect labels; 2) INB method; 3) NLNN method; 4) NAL method; 5) NB(correct), which is a NB model trained on the training set with correct labels.\\

\begin{figure}
\centering
\vskip -0.2in
\includegraphics[scale=0.6]{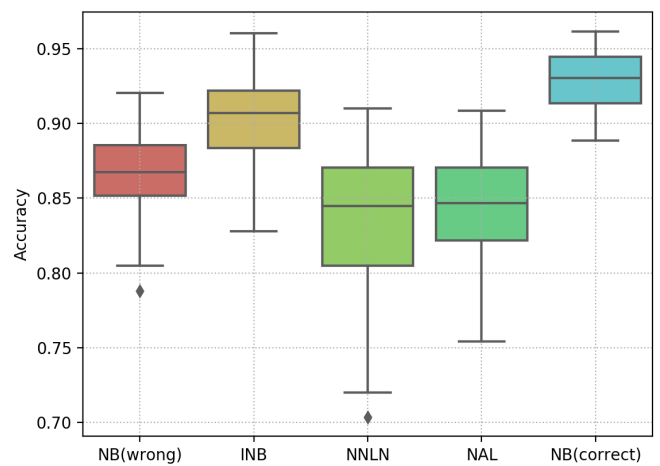}
\caption{Classification accuracy results on the \textit{Live Streaming Dialog Dataset}.}\label{fig:audi}
\vskip -0.2in
\end{figure}

\textbf{Experimental Results}. The mislabeling effect on the \textit{Live Streaming Dialog Dataset} is -7.32\%. Figure \ref{fig:audi} presents classification accuracy results obtained by the five methods in $B=100$ dataset splits. The proposed INB method outperforms the other methods. On average, by adding the mislabeling mechanism, the INB method has improved the classification accuracy of the NB method by 3.7\%. The prediction performances of the INB method are very close to that of the NB(correct) method, which is the NB method trained on the data whose labels are all correct.

\section{DISCUSSION}

We proposed an INB method for text classification. The INB method is analytically simple and free of subjective judgements on the correct and incorrect labels. By specifying the mechanism of generating incorrect labels, we optimize the corresponding log-likelihood function by an EM algorithm. We conduct simulation experiments to illustrate the advantage of the proposed INB method. Furthermore, we present a \textit{Live Streaming Dialog Dataset} to study the empirical performances of the INB method. Both numerical results suggest the INB method has competitive performances with mislabeled dataset.\\

To conclude this article, we discuss here an interesting topic for future study. In this work, we focus on the classical NB method with binary features. On the other hand, continuous variables are often encountered in real practice. Then, how to accommodate continuous features into the proposed INB framework becomes a problem of great interest. The key issue here is what kind of distribution assumptions should be imposed for the continuous features. One straightforward solution could be the Gaussian assumption. In this case, an EM-type algorithm can be readily developed (see Appendix \ref{Continuous Features} for details). However, whether this is the best assumption for optimal empirical performances is not immediately clear. More flexible nonparametric distribution assumptions might be promising alternatives. A more in-depth analysis should be definitely needed in this regard.

\appendix

\section*{Appendix}

\begin{table*}[t]
\caption{Finite sample performances of different methods with different $\rho_{kk}$ intervals and unbalanced sample sizes.}\label{simulation-table-unbalanced}
{\small
\begin{tabular}{@{}c|c|cc|ccccc|ccccc|c@{}}
\toprule
                                &       & \multicolumn{2}{c|}{MSE$\times 10^{-3}$} & \multicolumn{5}{c|}{ACC (\%)}    & \multicolumn{5}{c|}{AUC (\%)}    &                  \\
$\rho_{kk}$ Intervals           & $n$   & NB                  & INB                & NB   & INB  & NB-T & NLNN & NAL  & NB   & INB  & NB-T & NLNN & NAL  & $\Delta$ACC (\%) \\ \midrule
\multirow{3}{*}{{[}0.55, 0.65)} & 1000  & 2.5                 & 1.6                & 72.4 & 90.7 & 97.0 & 23.5 & 47.0 & 92.3 & 99.1 & 99.9 & 56.3 & 68.0 & -19.4            \\
                                & 5000  & 1.3                 & 0.3                & 88.9 & 95.2 & 96.2 & 38.1 & 78.8 & 98.5 & 99.6 & 99.8 & 64.2 & 94.4 & -6.4             \\
                                & 10000 & 1.2                 & 0.2                & 92.0 & 96.4 & 96.5 & 52.3 & 89.2 & 99.4 & 99.8 & 99.8 & 76.3 & 98.3 & -4.3             \\ \midrule
\multirow{3}{*}{{[}0.65, 0.75)} & 1000  & 2.1                 & 1.6                & 81.5 & 90.8 & 97.0 & 27.6 & 50.1 & 96.5 & 99.1 & 99.9 & 58.4 & 73.9 & -10.2            \\
                                & 5000  & 0.9                 & 0.3                & 92.3 & 95.2 & 96.2 & 44.0 & 84.3 & 99.3 & 99.6 & 99.8 & 66.8 & 96.7 & -3.0             \\
                                & 10000 & 0.8                 & 0.2                & 94.1 & 96.4 & 96.5 & 61.4 & 91.6 & 99.6 & 99.8 & 99.8 & 77.6 & 99.0 & -2.3             \\ \midrule
\multirow{3}{*}{{[}0.75, 0.85)} & 1000  & 1.8                 & 1.6                & 86.5 & 90.9 & 97.0 & 32.1 & 52.4 & 97.9 & 99.1 & 99.9 & 60.5 & 77.4 & -5.3             \\
                                & 5000  & 0.6                 & 0.3                & 94.0 & 95.2 & 96.2 & 51.1 & 87.7 & 99.5 & 99.6 & 99.8 & 68.7 & 97.9 & -1.3             \\
                                & 10000 & 0.5                 & 0.2                & 95.3 & 96.3 & 96.5 & 61.5 & 93.0 & 99.7 & 99.8 & 99.8 & 74.1 & 99.3 & -1.1             \\ \midrule
\multirow{3}{*}{{[}0.85, 0.95)} & 1000  & 1.5                 & 1.5                & 90.2 & 91.3 & 97.0 & 33.7 & 56.3 & 98.8 & 99.2 & 99.9 & 60.7 & 80.1 & -1.5             \\
                                & 5000  & 0.4                 & 0.3                & 94.8 & 95.2 & 96.2 & 47.6 & 89.4 & 99.6 & 99.6 & 99.8 & 66.4 & 98.4 & -0.4             \\
                                & 10000 & 0.2                 & 0.2                & 96.1 & 96.3 & 96.5 & 62.9 & 94.1 & 99.8 & 99.8 & 99.8 & 75.5 & 99.5 & -0.3             \\ \midrule
\multirow{3}{*}{{[}1.0, 1.0{]}} & 1000  & 1.4                 & 1.5                & 91.8 & 91.7 & 97.0 & 37.4 & 58.5 & 99.3 & 99.3 & 99.9 & 61.5 & 81.3 & 0.0              \\
                                & 5000  & 0.3                 & 0.3                & 95.3 & 95.3 & 96.2 & 47.4 & 91.4 & 99.6 & 99.6 & 99.8 & 64.6 & 98.9 & 0.0              \\
                                & 10000 & 0.1                 & 0.1                & 96.4 & 96.4 & 96.5 & 59.4 & 95.1 & 99.8 & 99.8 & 99.8 & 72.1 & 99.7 & 0.0              \\ \bottomrule
\end{tabular}
}
\end{table*}

\subsection{Continuous Features} \label{Continuous Features}

Here we consider how to add continuous features to the INB model with Gaussian assumptions. Suppose there are $d_1$ binary features and $d_2$ continuous features. Denote the binary features by  $X_i=(X_{i1}, \cdots, X_{id_1})^\top$. Denote the continuous features by $Z_i=(Z_{i1}, \cdots, Z_{id_2})^\top$. We collect all the $X_i$s into $\mathbb{X}$ and $Z_i$s into $\mathbb{Z}$. Similar to the Gaussian Na{\"i}ve Bayes model, we assume that $p(z_{ij}|Y_i^*=k) =  \phi_{jk}(z_{ij})$ for $1 \leq k \leq K$ and $1 \leq j \leq d_2$, where $\phi_{jk}(\cdot)$ is a normal distribution density function with mean $\mu_{jk}$ and standard deviation $\sigma_{jk}$. The whole parameter set becomes $\theta = \{p_{jk}, 1 \leq j \leq d_1, 1 \leq k \leq K \} \cup \{ \pi_k, 1 \leq k \leq K\} \cup \{ \rho_{k_1 k_2}, 1 \leq k_1, k_2 \leq K \} \cup \{ \mu_{jk}, 1 \leq j \leq d_2, 1 \leq k \leq K\} \cup \{ \sigma_{jk}, 1 \leq j \leq d_2, 1 \leq k \leq K\}$. Then the corresponding log-likelihood becomes
{\small
\begin{align}
    \ell_c(\theta) =& \ln{P(\mathbb{X}, \mathbb{Z}, \mathbb{Y}, \mathbb{Y}^*|\theta)} \nonumber \\
    =& \sum_{i=1}^N \ln{P(Y_i^*|\theta)} + \sum_{i=1}^N \ln{P(Y_i|Y_i^*, \theta)} \nonumber \\
    &+ \sum_{i=1}^N \sum_{j=1}^{d_1} \ln{P(X_{ij}|Y_i^*, \theta)} + \sum_{i=1}^N \sum_{j=1}^{d_2} \ln{P(Z_{ij}|Y_i^*, \theta)} \nonumber \\
    =& \sum_{i = 1}^N \ln{\pi_{Y_i^*}} + \sum_{i = 1}^N \ln{\rho_{Y_i Y_i^*}}  + \sum_{i = 1}^N   \sum_{j=1}^{d_1}  X_{ij} \ln{p_{jY_i^*}}  \nonumber \\
    &+ \sum_{i = 1}^N \sum_{j=1}^{d_1} (1 - X_{ij}) \ln{(1 - p_{jY_i^*})} + \sum_{i=1}^N \sum_{j=1}^{d_2} \ln{\phi_{j Y_i^*}(Z_{ij})}. \nonumber
\end{align} Correspondingly, we can develop an EM algorithm similar to that in the {Subsection \ref{EM algorithm}}.\\

\noindent\textbf{E Step}. In the $t$th iteration, we compute the expectation of the log-likelihood $\ell_c(\theta)$ given the observed data $(\mathbb{X}, \mathbb{Z}, \mathbb{Y})$ and the current parameter estimate $\widehat{\theta}^{(t)}$. The conditional expectation can be derived as
{\small
\begin{gather*}
    Q\Big(\theta, \widehat{\theta}^{(t)}\Big) = E \Big\{ \ell_c(\theta) \Big| \mathbb{X}, \mathbb{Z}, \mathbb{Y}, \widehat{\theta}^{(t)}\Big\} \nonumber \\
    = \! \sum_{i=1}^N \sum_{k=1}^K \widehat{\gamma}_{ik}^{(t)} \ln{\pi_k} \! + \! \sum_{i=1}^N \sum_{k=1}^K \widehat{\gamma}_{i k}^{(t)} \ln{\rho_{Y_i k}} \! + \! \sum_{i=1}^N \sum_{j=1}^{d_1} \sum_{k=1}^K X_{ij} \widehat{\gamma}_{ik}^{(t)} \ln{p_{jk}} \nonumber \\
    + \! \sum_{i=1}^N \! \sum_{j=1}^{d_1} \! \sum_{k=1}^K (1-X_{ij})\widehat{\gamma}_{ik}^{(t)} \ln{(1-p_{jk})}
    \!+\! \sum_{i=1}^N \! \sum_{j=1}^{d_2} \! \sum_{k=1}^K  \widehat{ \gamma}_{ik}^{(t)} \ln{\phi_{jk}}(Z_{ij}),
\end{gather*}}where $\widehat{\gamma}_{ik}^{(t)}$ denotes $P(Y_i^* = k|X_i, Z_i, Y_i, \widehat{\theta}^{(t)})$ for $1 \leq i \leq N$ and $1 \leq k \leq K$. To compute $\widehat{\gamma}_{ik}^{(t)}$, we first define $\widehat{\zeta}_{ik}^{(t)}$ as follows
{\small
\begin{align*}
    \widehat{\zeta}_{ik}^{(t)} = \widehat{\pi}_k^{(t)} \widehat{\rho}_{Y_i k}^{(t)} \bigg[\prod_{j=1}^{d_1} \widehat{p}_{jk}^{(t) X_{ij}} \Big\{1 - \widehat{p}_{jk}^{(t)}\Big\}^{1 - X_{ij}}\bigg] \times \bigg[ \prod_{j=1}^{d_2} \widehat{\phi}^{(t)}_{jk}(Z_{ij})\bigg],
\end{align*}}where $\widehat{\phi}^{(t)}_{jk}(Z_{jk}) = 1 / \sqrt{2 \pi \widehat{\sigma}^{(t) 2}_{jk}} \times \exp\{-(Z_{jk} - \widehat{\mu}^{(t)}_{jk}) / (2 \widehat{\sigma}^{(t) 2}_{jk})\}$. Then we compute $\widehat{\gamma}_{ik}^{(t)}=\widehat{\zeta}_{ik}^{(t)}/\sum_{k=1}^K \widehat{\zeta}_{ik}^{(t)}$.\\

\textbf{M step}. In this step, we maximize the conditional expectation $Q(\theta, \widehat{\theta}^{(t)})$ to get the new estimate $\widehat{\theta}^{(t+1)}$. Note that $\sum_{k=1}^K \pi_k = 1$ and $\sum_{k_1 =1} \rho_{k_1 k_2} = 1$ for $1 \leq k_2 \leq K$. Based on the Lagrange multiplier method, we can define the Lagrangian function as
{\small 
\begin{align*}
    Q = Q\left(\theta, \widehat{\theta}^{(t)}\right) + \lambda \left( 1 - \sum_{k=1}^K \pi_k \right) + \sum_{k_2=1}^K \lambda_{k_2} \left(1 -  \sum_{k_1=1}  \rho_{k_1 k_2}\right),
\end{align*}}where $\lambda$ and $\lambda_{k_2}$ for $1 \leq k_2 \leq K$ are additional parameters used in the Lagrange multiplier method. After maximizing the Lagrangian function $Q$, the updated parameters in the $(t+1)$th iteration are computed as
{\small
\begin{align*}
    & \widehat{\pi}_k^{(t+1)} = \sum_{i=1}^N \widehat{\gamma}_{ik}^{(t)}/ N , \ 1 \leq k \leq K, \nonumber \\
    & \widehat{p}_{jk}^{(t+1)} = \left(\sum_{i=1}^N X_{ij} \widehat{\gamma}_{ik}^{(t)}\right) \bigg/ \left( \sum_{i=1}^N \widehat{\gamma}_{ik}^{(t)}\right), \ 1 \leq j \leq d_1,
    \ 1 \leq k \leq K, \nonumber \\
    & \widehat{\rho}^{(t+1)}_{k_1 k_2} = \left(\sum_{i=1}^N I(Y_i = k_1) \widehat{\gamma}_{i k_2}^{(t)} \right) \bigg/ \left(\sum_{i=1}^N \widehat{\gamma}_{i k_2}^{(t)}\right), \ 1 \leq k_1, k_2 \leq K, \nonumber \\
    & \widehat{\mu}_{jk}^{(t+1)} = \left(\sum_{i=1}^N Z_{ij} \widehat{\gamma}_{ik}^{(t)}\right) \bigg/ \left(\sum_{i=1}^N \widehat{\gamma}_{ik}^{(t)}\right), \ 1 \leq j \leq d_2, \ 1 \leq k \leq K, \\
    & \Big(\widehat{\sigma}_{jk}^{(t+1)}\Big)^2 = \left\{\sum_{i=1}^N \widehat{\gamma}_{ik}^{(t)} \Big( Z_{ij} - \widehat{\mu}_{jk}^{(t+1)} \Big)^2 \right\} \bigg/ \left(\sum_{i=1}^N \widehat{\gamma}_{ik}^{(t)}\right).
\end{align*}}The EM algorithm designed for both binary and continuous features is similar to that in the {Subsection \ref{EM algorithm}}. Since we have continuous features $Z_i$s, updating $\widehat{\gamma}_{ik}^{(t)}$ involves in the parameters $(\widehat{\mu}_{jk}^{(t)}, \widehat{\sigma}_{jk}^{(t)})$ for $1 \leq j \leq d_2$ and $1 \leq k \leq K$. Furthermore, we also have to update $(\widehat{\mu}_{jk}^{(t)}, \widehat{\sigma}_{jk}^{(t)})$ in the M Step.
}

\subsection{Unbalanced Sample Size}\label{unbalanced sample size}

We next study the unbalanced sample size effect. To this end, we follow the same data generating strategy as Section \ref{simulation} but with only two modifications. 
The first modification is that the prior probability of the class 1 is set to be 3 times the probability of other classes. That is $\pi_1=0.428$ and $\pi_k=0.143$ for $k>1$. The second modification is that the data size is set to be $n=1000$, 5000, and 10000.
We set the classes to be $K=5$. For validation purpose, about 20\% of the simulated data are reserved for testing in each random replication. The detailed results are given in Table \ref{simulation-table-unbalanced}. We find that the results are qualitatively similar to that of Table \ref{simulation-table}.

\subsection{Flow Chart of the Labeling Process}\label{flowchart}

\begin{figure*}
\centering
\vskip -0.2in
\includegraphics[scale=0.3]{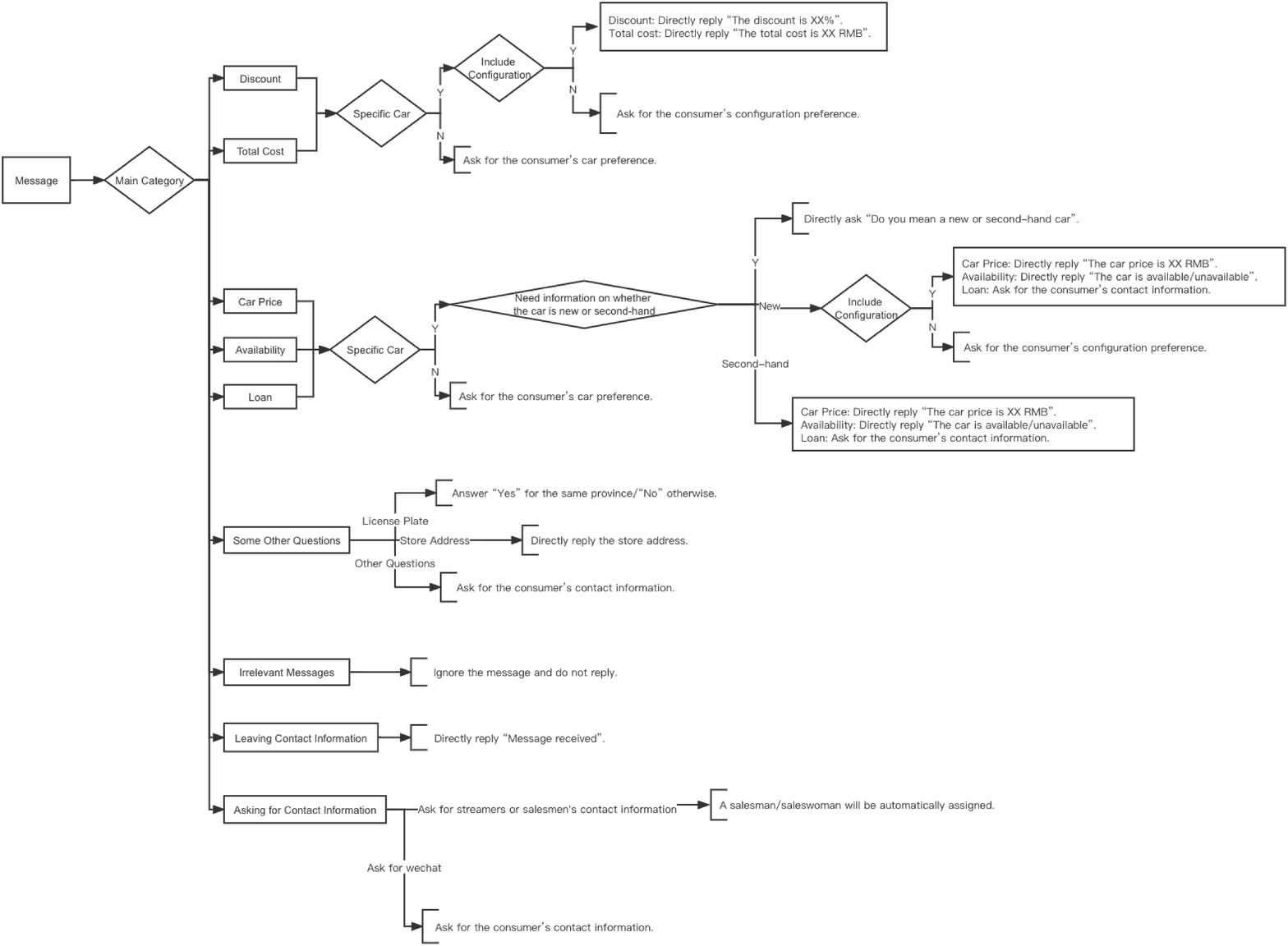}
\caption{Flow chart of the reply procedures.}\label{fig:flowchart}
\end{figure*}

To consumers' questions automatically, we first classify those messages into different categories according to a carefully designed reply strategy; see the \engordnumber{3} column in Table \ref{category}. Whenever a message is correctly classified into this category, it is then replied to the consumer according the procedures listed in Figure \ref{fig:flowchart}. For example, if the consumer asks about discounts about a specific car without configuration information, then he or she will receive a message asking for his or her configuration preference.

\section*{Acknowledgements}
The research of Xuening Zhu is supported by the National Natural Science Foundation of China (NSFC, 11901105, 71991472, U1811461), the Shanghai Sailing Program for Youth Science and Technology Excellence (19YF1402700), and the Fudan-Xinzailing Joint Research Centre for Big Data, School of Data Science, Fudan University.
Feifei Wang's research is supported by National Natural Science Foundation of China (72001205) and Chinese National Statistical Science Research Project (2022LD06).
Hansheng Wang's research is partially supported by National Natural Science Foundation of China (No. 11831008) and also  partially supported by the Open Research Fund of Key Laboratory of Advanced Theory and Application in Statistics and Data Science (KLATASDS-MOE-ECNU-KLATASDS2101).

\bibliographystyle{imsart-number}
\bibliography{reference}

\end{document}